\documentclass{article}

\PassOptionsToPackage{numbers, compress}{natbib}
\usepackage[preprint]{neurips_2026}

\usepackage[utf8]{inputenc} 
\usepackage[T1]{fontenc}    
\usepackage{hyperref}       
\usepackage{url}            
\usepackage{booktabs}       
\usepackage{amsfonts}       
\usepackage{nicefrac}       
\usepackage{microtype}      
\usepackage{xcolor}         

\usepackage{graphicx}
\usepackage{amssymb}
\usepackage{amsmath}
\usepackage{amsthm}
\usepackage{wrapfig}
\usepackage{subcaption}
\usepackage{enumitem}
\newtheorem{theorem}{Theorem}

\title{Signal-to-Noise Ratio and Sample Size Govern Representational Alignment in Neural Networks}

\author{
  Ali Hussaini Umar\thanks{Theoretical and Scientific Data Science (TSDS) group at the International School for Advanced Studies (SISSA).} \\
  SISSA\\
  Trieste, Italy \\
  \texttt{aumar@sissa.it} \\
  \And
  Alessandro Laio$^*$\thanks{Condensed Matter and Statistical Physics section at the International Centre for Theoretical Physics (ICTP).} \\
  SISSA $\&$ ICTP \\
  Trieste, Italy \\
  \texttt{laio@sissa.it}\\
}

\begin{document}

\maketitle

\begin{abstract}
  Neural networks are known to develop latent representations that are \emph{aligned}, namely structurally similar across networks trained with different architectures, training protocols, or training datasets. We study this phenomenon in a controlled setting, where we train an ensemble of networks on regression and classification tasks using training sets perturbed by independent realizations of a noise process.  We show that the signal-to-noise ratio (SNR) and the training sample size influence the alignment in qualitatively similar ways in networks trained on real-world datasets and in an extremely simple \emph{linear} network with a single hidden layer, for which the alignment can be estimated analytically. Across linear and nonlinear networks, regression and classification tasks, and both synthetic and real-world data, we consistently observe that alignment varies monotonically with SNR but non-monotonically with training sample size. In particular, the alignment is minimized near the interpolation threshold, and a stronger alignment does not necessarily correspond to better generalization error. These findings reveal a non-trivial dependence of alignment on data quality and quantity, decoupled from generalization performance.
\end{abstract}

\section{Introduction}

Deep learning has achieved remarkable empirical success across a wide range of domains, including computer vision~\cite{lecun2015deep, krizhevsky2012imagenet}, natural language processing~\cite{cho2014learning,vaswani2017attention}, and speech recognition \cite{hinton2012deep}. A common hypothesis attributes this success to neural networks' ability to learn meaningful internal representations of data~\cite{bengio2013representation, lecun2015deep, goodfellow2016deep}. This perspective has motivated a growing body of work aimed at quantifying, understanding, and predicting representational alignment~\cite{sucholutsky2025getting}. In particular, the mechanism by which independently trained networks differ in initialization, architecture, training objectives, or training datasets converges to similar internal representations~\cite{kriegeskorte2008representational, raghu2017svcca, pmlr-v97-kornblith19a,bansal2021revisiting,huh2024platonic, acevedo2025quantitative,groger2026revisiting}.

Despite significant empirical advances, a theoretical and mechanistic understanding of such a mechanism remains elusive. Most existing studies focus on quantifying alignment -- using tools such as representational similarity analysis (RSA)~\cite{kriegeskorte2008representational}, singular vector canonical correlation analysis (SVCCA)~\cite{raghu2017svcca}, centered kernel alignment (CKA)~\cite{pmlr-v97-kornblith19a}, generalized shape metrics~\cite{williams2021generalized}, and nearest-neighbor metrics~\cite{klabunde2025similarity,huh2024platonic} -- but leave several fundamental questions open: \textit{(i)} how alignment depends on the quality and quantity of training data; \textit{(ii)} whether the mechanism driving alignment differs between the underparameterized and overparameterized regimes; and \textit{(iii)} whether representational alignment is predictive of the quality of the model, measured by generalization error. Addressing these questions would be of help for developing a principled theory of representation learning and for characterizing the conditions under which independently trained networks encode the same information.

We first explore these questions in a setting where the learning problem is sufficiently simple to admit exact analysis, yet rich enough to exhibit the development of nontrivial alignment phenomena in the hidden representations. With this goal in mind, we consider a two-layer linear network, for which we are able to quantify the properties of the hidden representation by a rigorous analytic treatment. Although deep linear networks ultimately implement a linear transformation of the input, they do so through a layered, compositional process that gives rise to intermediate hidden representations~\cite{saxe2013exact}. Moreover, their loss landscape is non-convex, and the set of global minima forms a high-dimensional solution manifold whose geometry is shaped by both the statistics of the training data and the network architecture~\cite{baldi1989neural, saxe2013exact}. These properties make linear networks a useful surrogate model for investigating representation learning~\cite{saxe2019mathematical, braun2025not, domine2025from}. We then extend our analysis to nonlinear networks via numerical simulation, on the same tasks using the linear network, and on different tasks using real-world datasets, with the aim of identifying similarities and differences with respect to the linear setting.

Our analysis focuses on two factors that can affect hidden representations: the noise in the training set, and the training sample size. The effect of those two factors can be quantified rigorously by analytic tools in simple networks and through extensive numerical analysis in generic networks.

Since our analysis extends to networks in which the representation manifold is highly curved and nonlinear~\cite{ansuini2019intrinsic}, we do not use alignment measures such as CKA~\cite{pmlr-v97-kornblith19a} and SVCCA~\cite{raghu2017svcca}, which are designed to capture correspondences expressible by a linear transformation and are therefore ill-suited for such settings (see related work section Sec.~\ref{sec_related_work} for more reasons why such measures are not suited).

In this work, we say that two representations are aligned if the knowledge of the neighbors in one representation allows predicting the neighbors in the other~\cite{huh2024platonic,acevedo2025quantitative,klabunde2025similarity}. We quantify this predictability by a novel statistical measure inspired by the Information Imbalance (II)~\cite{glielmo2022ranking}, defined as the entropy of the rank distribution in the target representation conditioned on the observation that the rank in the source representation is small~\cite{del2024robust}. This quantity, which we call \textit{Conditional Copula Entropy} (CCE), at variance with the II, can be estimated analytically in deep linear networks. The availability of a reference system (deep linear network) on which our statistical measure can be computed analytically and compared with simulations provides a robust theoretical foundation for our analysis. 

Our main contributions are summarized as follows:
\begin{itemize}
    \item  We develop a theoretical framework for analyzing representational alignment using the conditional distribution of copula variables and derive closed-form expressions for alignment in two-layer linear neural networks.
    \item We show analytically in linear networks and numerically in nonlinear networks, that alignment depends on signal-to-noise ratio (SNR) and training sample size across regression and classification tasks with real-world datasets. In all the cases we considered, alignment increases monotonically with SNR but varies non-monotonically with training sample size, attaining a minimum near the interpolation threshold, akin to the behavior observed in the generalization error. For the classification task, label noise plays the role of an inverse SNR.
    
    \item We show that better generalization performances does not necessarily imply more aligned internal representations. This decoupling is proved analytically in the linear network and numerically in the other cases we considered.
    
    \item Finally, we show that task performance is an insufficient proxy for representation quality; a network can achieve higher task performance, and yet learn less informative internal representations than a network with lower performance, highlighting the limitations of using performance metrics as a proxy for representation quality.
\end{itemize}
In general, our results provide theoretical and numerical insights that contribute toward building a solid theoretical foundation for representation learning, as well as identifying the conditions under which independently trained networks end up forming similarly structured internal representations.
\section{Representational Alignment from the Conditional Entropy of Rank Distribution}
\label{sec:II}
\paragraph{Information Imbalance.} The statistic we use to quantify the alignment between representations is directly related to the Information Imbalance (II) \citep{glielmo2022ranking},
which quantifies the relative information content of one representation space with respect to another by comparing how the distance ranks change between the two representations. For $N$ data points, the II from space $A$ to space $B$ is
\begin{equation}
   \label{II}
    \Delta(A\rightarrow B) = \frac{2}{N(N-1)} \sum_{i,j \mid r_{i,j}^A = 1} r_{i,j}^B,
\end{equation}
where $r_{i,j}^A \in \{1,\dots, N-1\}$ denotes the rank of the distance between points $i$
and $j$ in space $A$, ordered from nearest to farthest from $i$; $r_{i,j}^B$ is defined
analogously. Eq.~\ref{II} thus captures how well proximity in $A$ predicts proximity
in $B$. Values near zero indicate strong alignment, while $\Delta(A\to B)\approx 1$ indicates
that $A$ carries no information about $B$. The measure is generally asymmetric:
$\Delta(A\to B) < \Delta(B\to A)$ implies that $A$ is more informative than $B$.

\paragraph{From II to Conditional Copula Entropy.} 
In this work, we use an information-theoretic quantity related to the II, which can be computed analytically in some data scenarios of relevance for our analysis. In ref. \cite{del2024robust}, it is shown that II provides an upper bound of a function of the conditional entropy of the copula variables associated with the pairwise distance variables
in spaces $A$ and $B$:
\begin{equation}
    \label{inequality_main}
    \Delta(A \rightarrow B) \gtrsim 2 \exp\!\left( \lim_{\eta \rightarrow 0} H(c_B\mid c_A = \eta)  - 1 \right),
\end{equation}
where $c_A$ and $c_B$ are the copula variables.
The negative of the entropy\footnote{$
    -\lim_{\eta \rightarrow 0} H(c_B\mid c_A = \eta)=\lim_{\eta \to 0} D_{\mathrm{KL}}\big[p(c_B \mid c_A = \eta) \,\|\, p(c_B)\big]$} term measures how much knowing that two points
are infinitesimally close in $A$ reduces uncertainty about their distance in $B$ -- a geometric
characterization of local representational alignment via information-theoretic quantity. Motivated by this interpretation, we measure representational alignment by the entropy entering the relation in \ref{inequality_main}, which, for brevity, we call \textit{Conditional Copula Entropy} (CCE):
\begin{equation}
\label{CCE}
    \text{CCE}(A \to B) = -\lim_{\eta \to 0}  H(c_B\mid c_A = \eta).
\end{equation}
 If $\text{CCE}(A \to B) = \text{CCE}(B \to A) = 0$, the two
representations are geometrically independent and thus misaligned. Positive values indicate
that the neighborhood structure of one space is informative about the other, with larger values reflecting stronger alignment. Like the II, the CCE can be asymmetric:
$\text{CCE}(A \to B) > \text{CCE}(B \to A)$ implies that $A$ is more informative about $B$ than vice versa. For more about CCE check Sec.~\ref{sec_about_CCE} in the Appendix.
\paragraph{Estimating the CCE.}
In practice, the conditional distribution $p(c_B \mid c_A)$ is unknown and must be estimated
from data. Let $d_A$ and $d_B$ denote the pairwise distance random variables in spaces $A$ and $B$. The copulas variables associated with $d_A$ and $d_B$ are defined as
\begin{equation*}
    (c_A, c_B) = (F_A(d_A),\, F_B(d_B)),
\end{equation*}
with joint density $p(c_A, c_B)$ where $F_A$ and $F_B$ are the cumulative distribution functions of $d_A$ and $d_B$,
respectively. Since the marginals of $c_A$ and $c_B$ are uniform on $[0,1]$~\citep{casella2024statistical}, the copula variable $c_A(i,j)$ can be estimated empirically as $r^A_{i,j}/N$ (and analogously for $c_B$). Consequently, distance ranks and copula
variables are statistically identical; this relation allows us to estimate the CCE directly from the conditional rank samples. Therefore, given a set of $N$ conditional ranks
\begin{equation*}
       R_{B \mid A} = \bigl\{ r^B_{i,j} \in [1, N-1] : r^A_{i,j} = 1 \bigr\},
\end{equation*}
we partition the rank support $[1, N-1]$ into $M \approx \sqrt{N}$ equal bins and estimate the entropy via
the histogram-based estimator of ranks in $R_{B \mid A}$ \citep{beirlant1997nonparametric, kraskov2004estimating}. Letting $p_i$ denote the
empirical density in bin $i$, CCE (Eq.~\ref{CCE}) is approximated as
\begin{equation}
\label{ECCE}
    \text{CCE}(A \to B) \approx \log M + \sum_{i=1}^M p_i \log p_i.
\end{equation}
We find that Eq.~\ref{ECCE} provides a good estimate of the true entropy for sufficiently large $N$, and we use it to compute CCE numerically~\cite{beirlant1997nonparametric}.

\paragraph{CCE for one-dimensional embedding spaces.}
In ref. \cite{umar2026effect}, it is shown that a closed-form expression for the CCE can be derived when $A$ and $B$ are one-dimensional projections of a Gaussian random vector. This result is as follows:
given $x \sim \mathcal{N}(0, I_d)$, and define $A = u^\top x$ and $B = v^\top x$, where $u, v \in \mathbb{R}^d$. Then $A$ and $B$ are jointly Gaussian, and the CCE admits the following closed-form expression
\begin{equation} 
\label{CCE_form}
    \text{CCE}(A \to B) = \text{CCE}(B \to A) =-\frac{1}{2}\log(1 - \rho^2) - \frac{1}{2}\rho^2,
\end{equation}
where $\rho = \mathbb{E}\big[u^\top v\big]/
\sqrt{
\mathbb{E}\big[\|u\|^2\big]\;
\mathbb{E}\big[\|v\|^2\big]
}$ denotes the correlation coefficient between $A$ and $B$.
In this work, we leverage this result to derive the closed-form expression for the CCE between the representations of independently trained linear networks.
\section{Linear Teacher-Student Setting}
In this work, we analyze the alignment between representations of independently trained networks in different settings (a regression task on synthetic data and a classification task on real data). First,
we consider a setting in which a rigorous analytic treatment is possible; this allows us, among other reasons, to verify that our alignment estimator agrees with theoretical predictions.
Following Ref.~\cite{hastie2022surprises}, we consider a supervised learning problem based on data generated via a noisy linear teacher of the form:
\begin{equation}
\label{teacher_model}
    y_i = w^* \cdot x_i + \epsilon_i,
\end{equation}
where $x_i \in \mathbb{R}^{d}$ are input features, $w^* \in \mathbb{R}^{d}$ is the teacher weight vector and $\epsilon_i\in \mathbb{R}$ is additive Gaussian noise. The inputs are drawn independently and identically distributed (i.i.d.)\ from $\mathcal{N}(0, I_d/d)$. The entries of $w^*$ and $\epsilon_i$ are i.i.d. drawn from $\mathcal{N}(0, \sigma_w^2)$ and  $\mathcal{N}(0, \sigma_\epsilon^2)$, respectively. We define the $\text{SNR}= \sigma_w^2 / \sigma_\epsilon^2$.

Using the setup in Ref.~\cite{advani2020high}, the student is a two-layer linear network with input-output mapping
\begin{equation}
\label{student_dln}
    \hat{y}_i = W_2 W_1 x_i =: W_{\mathrm{tot}} x_i,
\end{equation}
where $W_1 \in \mathbb{R}^{k \times d}$ and $W_2 \in \mathbb{R}^{1\times k}$. The network is trained on $n$ samples $\mathcal{D}=\{(x_i,y_i)\}_{i\le n}$ generated from Eq.~\ref{teacher_model} using full-batch gradient descent to minimize the mean squared error (MSE)
\begin{equation}
\label{loss}
    \mathcal{L}(W_1, W_2) = \frac{1}{n} \sum_{i=1}^n (\hat{y}_i - y_i)^2 = \frac{1}{n} \| W_{\mathrm{tot}} X^\top - y \|^2,
\end{equation}
where $X \in \mathbb{R}^{n \times d}$ is the data matrix and $y \in \mathbb{R}^{n}$ the vector of the corresponding responses. 
Here, we shift focus from studying the student's performance on unseen data to understanding how the internal representations of identical student networks -- each trained by gradient descent on independent datasets generated by the same teacher using independent realizations of the inputs and noise -- align with one another.
Specifically, our objective is to characterize how CCE depends on SNR and the number of training data $n$. In this work, we study only the learned representation at the global minimum of the loss 
; we leave the study of the dynamics for future work. 

\paragraph{Gradient flow dynamics and fixed points.}
Under the MSE loss (Eq.~\ref{loss}), the gradient flow dynamics of the two-layer
linear network (Eq.\ref{student_dln}) weights are:
\begin{align}
    \label{main_gradient_flow_W1}
    \tau \frac{dW_1}{dt} &= W_2^\top \bigl(yX - W_2 W_1 X^\top X\bigr), \\
    \label{main_gradient_flow_W2}
    \tau \frac{dW_2}{dt} &= \bigl(yX - W_2 W_1 X^\top X\bigr) W_1^\top,
\end{align}
where $\tau = n/2$ is a time constant. 
Equations \ref{main_gradient_flow_W1}--\ref{main_gradient_flow_W2} admit a manifold of
stable fixed points~\cite{baldi1989neural, saxe2013exact}:
\begin{equation*}
    \{W_1, W_2 \mid W_2 W_1 X^\top X = yX\}.
\end{equation*}
We work in the small-initialization (rich) regime~\citep{woodworth2020kernel}, and adopt the low-rank parameterization of~\citet{advani2020high}, which
exploits the singular value decomposition (SVD) of the input covariance. Specifically, we set $
    W_1 = r_2\, z V^\top, \; W_2 = u\, r_2^\top,$
where $V$ contains the right singular vectors of $X$,
$z \in \mathbb{R}^{d}$ encodes alignment with the input principal
components, $r_2 \in \mathbb{R}^{k}$ is a unit-norm vector capturing
representational degrees of freedom, and $u \in \mathbb{R}$ is a scalar.
With this parameterization, $u$ and $z$ are the only variational parameters. Setting $v = uz$, at the global minimum
\begin{equation}
   \label{main_v_star}
   v^* = w^* V + \tilde{\epsilon}\,\Lambda^{-1/2}
\end{equation}
where $\Lambda$ is the diagonal matrix of eigenvalues of $X^\top X$ and
$\tilde{\epsilon} \sim \mathcal{N}(0, \sigma_\epsilon I_d)$.

\section{Results}
\label{sec_results}
We present our results, organized as follows. We begin by presenting a closed-form expression for the CCE between the representations of linear networks trained on independent realizations of the inputs and noise, followed by simulation results that validate the theory. We then demonstrate that qualitatively similar behavior emerges in nonlinear networks, and show that these findings extend to real-world classification tasks. All simulations were conducted on NVIDIA RTX 6000 Ada Generation GPUs.

\subsection{Representational alignment in Deep Linear Network}
\begin{figure}[h]
    \centering
    \vspace{-8pt}
    \includegraphics[width=0.9\linewidth]{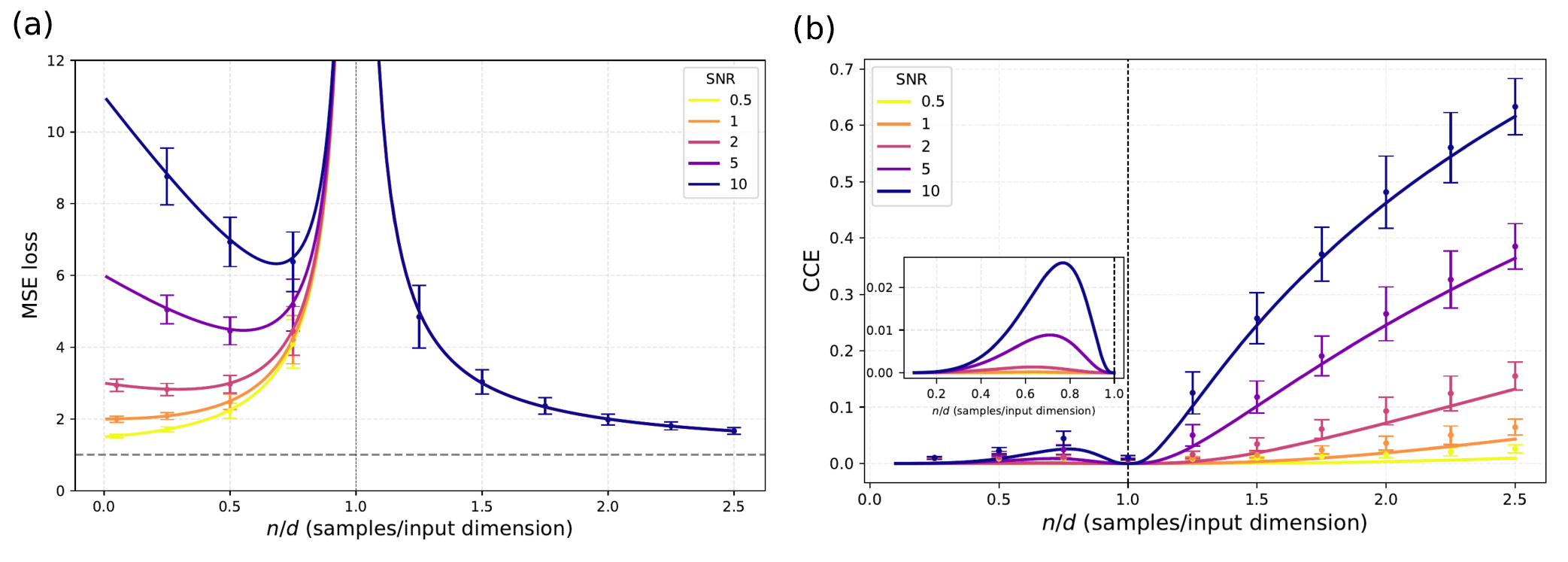}
     \vspace{-8pt} 
   \caption{\textbf{Representational alignment depends on the signal-to-noise (SNR) ratio and increases non-monotonically with training sample size (n)}. Panel (a): presents the generalization error of a two-layer linear network as a function of training sample size divided by the network's parameters for different SNR (at global minimum). The solid lines correspond to the theoretical predictions, while the dots with bars denote average results from numerical simulation. Panel (b): presents the  Conditional Copula Entropy (CCE) between hidden representations of identical networks trained on independent datasets for different SNR. The solid line indicates the theoretical CCE, while dots with bars indicate the empirical CCE. 
   Parameters: $d=200$ (input dimension), $k=100$ (hidden units),  $n_{test} =10^4$ (test samples).}
    \vspace{-10pt} 
    \label{alignment_dln}
\end{figure}
We estimate the asymptotic alignment using the following theorem.  
\begin{theorem}[Asymptotic Representational Alignment from Independent Training]
\label{thrm2}
Consider two deep linear networks
\[
\hat{y}_i^a = W_2^a W_1^a x_i, \qquad \hat{y}_i^b = W_2^b W_1^b x_i,
\]
trained independently by gradient descent from small random initializations on independent datasets $\mathcal{D}^a = \{X^a, y^a\}$ and $\mathcal{D}^b = \{X^b, y^b\}$, drawn independently from the data distribution \eqref{teacher_model}. Suppose the input covariance matrices are simultaneously diagonalizable with a shared eigenbasis,
\[
X^{a\top} X^a = V \Lambda^a V^\top, \qquad X^{b\top} X^b = V \Lambda^b V^\top,
\]
and that the first-layer weights admit the aligned low-rank parameterization
\[
W_1^a = r^a z^a V^\top, \qquad W_1^b = r^b z^b V^\top,
\]
where $r^a, r^b$ are unit-norm vectors. As training time $t \to \infty$, gradient descent converges to a global minimum, and the asymptotic correlation between $z^a$ and $z^b$ satisfies
\begin{equation}
\label{asymptotic_alignment}
    \rho^* = 
    \begin{cases}
    \displaystyle
    \frac{\text{SNR}}{\text{SNR} + \frac{1}{\alpha - 1}}, & \alpha > 1, \\[12pt]
    \displaystyle
    \frac{\alpha \cdot\text{SNR}}{\text{SNR} + \frac{1}{1 - \alpha}}, & \alpha < 1,
    \end{cases}
    \end{equation}
where $\rho^* = \lim_{t \to \infty} \rho(t)$, $\mathrm{SNR} = \sigma_w^2 / \sigma_\epsilon^2$, and $\alpha = n/d$ denotes the ratio of training samples to input dimension.
\end{theorem}

The full derivation of this Theorem is presented in the Appendix (see Sec.~\ref{asymptotic_alignment_proof}). In summary, the result in Eq.~\ref{asymptotic_alignment} follows from the global minimum structure of the solution (Eq.~\ref{main_v_star}). At convergence, $z^{*a}$ takes the form
\begin{equation}
z^{*a} = \frac{1}{u^*}\!\left(w^* V + \tilde{\epsilon}^a (\Lambda^a)^{-1/2}\right),
\end{equation}
where $\Lambda^a$ is the diagonal matrix of eigenvalues of $X^{a\top} X^a$, $\tilde{\epsilon}^a \sim \mathcal{N}(0, \sigma_\epsilon^2 I_d)$, and $u^*$ is the magnitude of the learned weight. An analogous expression holds for $z^{*b}$. The correlation between the two representations is then
\begin{equation*} 
\rho^*
=
\frac{
\mathbb{E}\big[z^{*a} z^{*b\top}\big]
}{
\sqrt{
\mathbb{E}\big[\|z^{*a}\|^2\big]\;
\mathbb{E}\big[\|z^{*b}\|^2\big]
}
}
= \frac{\sum_{i=1}^{R} \sigma_w^2}{\sqrt{\bigg[\sum_{i=1}^{R} \left(\sigma_w^2 + \frac{\sigma_\epsilon^2}{\lambda_i^a}\right)\bigg] \bigg[\sum_{i=1}^{R} \left(\sigma_w^2 + \frac{\sigma_\epsilon^2}{\lambda_i^b}\right)\bigg]}},
\end{equation*}
where $\{\lambda_i^a\}$ and $\{\lambda_i^b\}$ are the non-zero eigenvalues of $X^{a\top} X^a$ and $X^{b\top} X^b$, respectively, $R$ is the rank of the input covariance matrix, and $\sigma_w^2$, $\sigma_\epsilon^2$ are the variances of the teacher weights and noise. In the proportional high-dimensional limit $n, d \to \infty$ with $\alpha = n/d$ fixed, the empirical spectral distribution of $X^\top X$ converges to the Marchenko--Pastur law~\citep{marvcenko1967distribution, bai2010spectral},
\begin{equation}
\label{mp_distn}
f(\lambda) = \frac{\sqrt{(\lambda_+ - \lambda)(\lambda - \lambda_-)}}{2\pi\lambda} + (1 - \alpha)^+ \delta(\lambda),
\end{equation}
where $\lambda_\pm = (1 \pm \sqrt{\alpha})^2$ and $(\cdot)^+ = \max(\cdot, 0)$. The first term describes the continuous bulk supported on $[\lambda_-, \lambda_+]$; the second captures the point mass at zero that arises when $\alpha < 1$. Replacing the empirical average over eigenvalues with an integral against $f$ yields the closed-form result in Eq.~\ref{asymptotic_alignment}.
 
By means of Theorem~\ref{thrm2}, we can now compute the degree of alignment between representations of independently trained deep linear networks via the CCE. Let $A = W_1^a X_{\mathrm{test}}^\top$ and $B = W_1^b X_{\mathrm{test}}^\top$ denote the hidden representations obtained from the test set. In terms of pairwise distances, $A$ and $B$ are equivalent to one-dimensional projections $\tilde{A} = \tilde{z}^a X_{\mathrm{test}}^\top$ and $\tilde{B} = \tilde{z}^b X_{\mathrm{test}}^\top$. Applying the result in Eq.~\ref{CCE_form}, the CCE has the following closed-form expression
\begin{equation}
\label{CCE_expression}
\mathrm{CCE}(A \to B) = -\frac{1}{2}\log\!\left(1 - \rho^{*2}\right) - \frac{1}{2}\rho^{*2},
\end{equation}
with $\rho^*$ given by Eq. \ref{asymptotic_alignment}.

It is worth noting that this result relies on two key assumptions. First, we assume that the input correlation matrices $X^{a\top} X^a$ and $X^{b\top} X^b$ share the same orthogonal singular vector matrix $V$. This assumption is common in other theoretical analyses of deep linear networks \cite{saxe2013exact,lampinen2018analytic,saxe2019mathematical,advani2020high,atanasov2022neural,domine2024lazy,jarvis2025theory}.
Second, we assume that the learned first-layer weights have a low-rank structure. It is a well-known result from prior work that deep linear networks tend to extract low-rank structures when initialized with small weights \cite{arora2019implicit,saxe2013exact}, which is a form of implicit inductive bias by the training algorithm \cite{kalimeris2019sgd,refinetti2023neural}. Both these assumptions will be validated by comparing the theory with the numerical simulation. 

Fig.~\ref{alignment_dln} (Panel~(a)) shows the average generalization error of a two-layer linear network
as a function of the training sample size divided by the input dimension $\alpha = n/d$. The solid curve corresponds to the theoretical prediction (see Sec.~\ref{exact_gen_error_dln} in Appendix for its explicit form), while the dots with bars correspond to the numerical simulation results, averages over $50$ independent experiments. The networks are evaluated after convergence, and we set $d=200$ and $k=100$. The empirical results agree with the theoretical curves. In particular, we recover the characteristic generalization behavior observed in shallow linear models reported by~\cite{krogh1992generalization,hastie2022surprises,advani2020high}, including the double descent phenomenon. The interpolation threshold occurs at $n = d$, reflecting that the network is functionally equivalent to a linear model with $d$ effective parameters.

We now shift to our main focus: understanding the relationship between internal representations learned by identical networks trained on independent datasets. Let $A$ and $B$ denote the hidden representations of two such networks evaluated on a common test input $X_{\text{test}}$. Eq.~\ref{CCE_expression} characterizes the directional alignment measures $\text{CCE}(A \to B)$ and $\text{CCE}(B \to A)$, which are symmetric in this case, and thus we only report one. Fig.~\ref{alignment_dln} (Panel~(b)) shows $\text{CCE}(A \to B)$ as a function of $\alpha$ for different SNR. We observe a strong dependence on SNR both in the underparameterized $(\alpha >1)$ and overparameterized $(\alpha <1)$ regimes, with alignment (i.e., higher $\text{CCE}(A \to B)$) improving as \text{SNR} increases. For any finite $\text{SNR} > 0$, $\text{CCE}(A \to B)$ exhibits a non-monotonic dependence on $\alpha$, mirroring the double descent behavior of the generalization error: it vanishes at the interpolation threshold $\alpha = 1$ and increases as $\alpha \to \infty$, with a \text{SNR}-dependent rate. 
At the interpolation threshold,  $\text{CCE}(A \to B)\approx 0$, regardless of the SNR. This value of CCE indicates essentially no alignment between $A$ and $B$. This is consistent with the divergence of the generalization error at interpolation.
Overall, representation alignment is governed by the SNR and varies non-monotonically with training sample size. 

Remarkably, the SNR influences representational alignment and generalization errors in different manners. 
In the underparameterized regime ($\alpha > 1$), networks trained with different values of SNR achieve the same generalization error, yet their internal representations differ: in this regime, a larger SNR leads to stronger alignment despite identical performance. In contrast, in the overparameterized regime ($\alpha < 1$), a larger SNR degrades the performance as $\alpha \to 0$, even though the corresponding representations are more strongly aligned than those of networks trained with a smaller SNR, which achieve better generalization in this regime.


\subsection{Alignment in Non-Linear Networks}
\begin{figure}[h]
    \centering
    \vspace{-8pt}
    \includegraphics[width=0.9\linewidth]{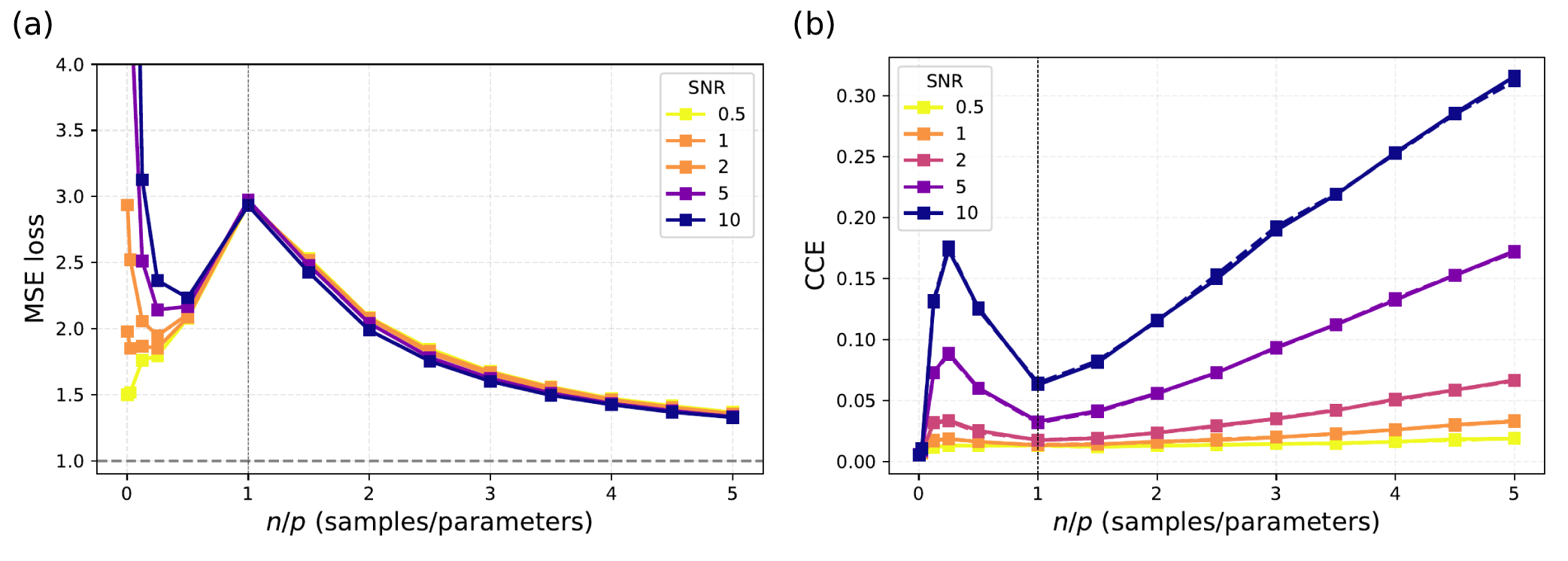}
     \vspace{-8pt}
   \caption{\textbf{Empirical results for non-linear neural network}.
    Panel (a): presents the average generalization error of a two-layer network with ReLU activation as a function of training sample size divided by the network's parameters for different SNR values. Panel (b): presents the average  Conditional Copula Entropy between hidden representations of identical networks trained on independent datasets. 
    Numerical results are averaged over 50 independent experiments. 
Parameters: $d=200$ (input dimension), $k=20$ (number of hidden units), $n_{test} = 10^4$ (test samples).}
    \label{alignment_drn}
     \vspace{-15pt}
\end{figure}

In this section, we extend our analysis to nonlinear networks by introducing a ReLU activation within the same architectural setting previously studied for linear models. The data-generative model and the training procedure remain unchanged. The presence of nonlinearity makes obtaining analytical results for CCE significantly more challenging. We therefore study the representations numerically. Fig.~\ref{alignment_drn} presents the results. Panel (a) shows the generalization error of ReLU networks for various values of the SNR, while Panel (b) reports the CCE between representations of networks trained on independent datasets with common SNR. Both quantities are plotted as a function of the ratio between the number of training samples $n$ and the number of network parameters $p = d \times k$. The curves are averaged over $50$ independent experiments, with $d = 200$ and $k = 20$.

Let $\gamma = n/p$. The generalization error exhibits a double descent phenomenon~\cite{belkin2019reconciling}, with a peak around $\gamma \approx 1$. For $\gamma > 1$, the error is largely independent of the SNR, whereas in the regime $\gamma \to 0$, the networks trained with smaller SNR perform better, consistent with the results for linear networks. Similarly, CCE between independently trained networks displays non-monotonic behavior,  with a sharp minimum near the same value where the generalization reached it peak ($\gamma \approx 1$). In both the underparameterized ($\gamma > 1$) and overparameterized ($\gamma < 1$) regimes, the CCE varies significantly with the SNR, with larger values of SNR leading to better alignment between representations. However, unlike the linear case, the generalization error does not diverge at the interpolation threshold, and the CCE does not approach zero as in the linear case. Instead, CCE converges to an SNR-dependent value, suggesting that even at the threshold, the learned representations share some information. Besides this difference, the qualitative behavior of both the generalization error and CCE is akin to that observed in the linear setting.

\begin{figure}[h]
    \centering
    \vspace{-8pt}
    \includegraphics[width=0.9\linewidth]{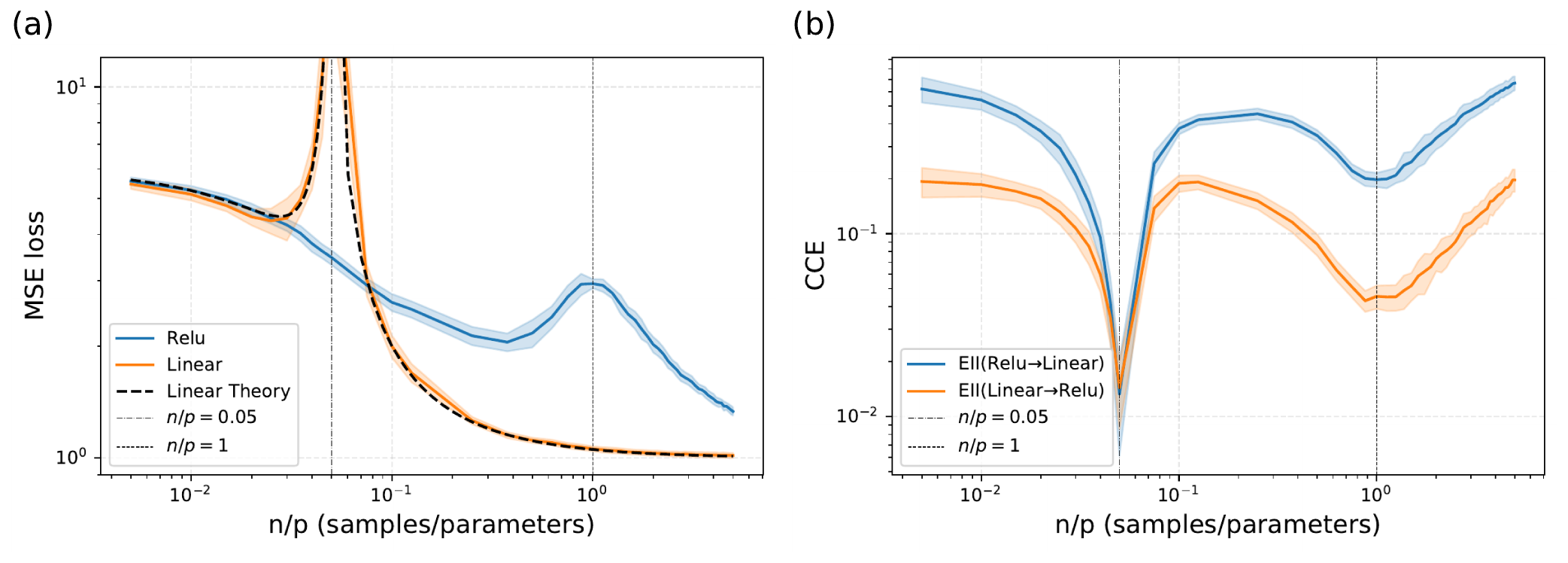}
     \vspace{-8pt}
   \caption{\textbf{Alignment between Representations of Linear and Nonlinear Network}.
    Panel (a): presents the average generalization error of a two-layer network with linear (orange curve) and ReLU (blue curve) activation as a function of training sample size divided by the network's parameters for SNR$=5$. Panel (b): presents the average  Conditional Copula Entropy between hidden representations of networks with Linear and ReLU activation trained on the same datasets. Orange curves represent CCE(Linear $\to$ ReLU), and blue curves represent CCE(ReLU $\to$ Linear). 
    Numerical results are averaged over 20 independent experiments. Parameters: $d=200$ (input dimension), $k=20$ (number of hidden units), $n_{test} = 10^4$ (test samples).}
    \label{alignment_dln_drn}
     \vspace{-15pt}
\end{figure}
Next, we compare the representations learned by the two networks we considered so far — one with linear activation and one with ReLU activation — trained on the same dataset ($d = 200$, $k = 20$, and SNR$=5$). Fig.~\ref{alignment_dln_drn}~(a) shows the generalization error as a function of $\gamma$. The two networks exhibit markedly different behaviors: the linear network (orange) consistently achieves a lower generalization error than the ReLU network (blue), except at the linear network's interpolation threshold ($n/d\times k = 0.05$), where the error diverged~\cite{hastie2022surprises,advani2020high}. This result is a direct consequence of the structure of the learning problem — by construction, the optimal student in this setting is linear~\cite{bishop2006pattern}, so the linear network operates closer to the Bayes-optimal learner~\cite{bishop2006pattern,cui2023bayes,cui2025bayes}.

Strikingly, however, as shown in panel (b) of Fig.~\ref{alignment_dln_drn}, the ReLU network learns internal representations that are \emph{more informative} — as measured by the CCE — than those of the linear network, despite the latter achieving a better generalization error.  At $n/d\times k = 0.05$, the CCE collapses to zero, due to the lack of structure in the learned representation of the linear network, making the two representations mutually unpredictable. The CCE also exhibit a local minimum at the interpolation threshold of the ReLU network ($n/d\times k = 1$), where the network interpolates the noisy training data~\cite{belkin2019reconciling,zhang2016understanding, mei2022generalization}. Notably, even at this threshold,  the representations learned by the ReLU network remain more informative than those of the linear network. This behavior is likely due to the well-known greater expressivity of nonlinear models compared to linear ones~\cite{barbier2025statistical}, which enables them to capture richer and more structured representations.



\subsection{Beyond synthetic settings: classification on real-world data}
\begin{figure}[h]
    \centering
    \vspace{-8pt}
    \includegraphics[width=0.9\linewidth]{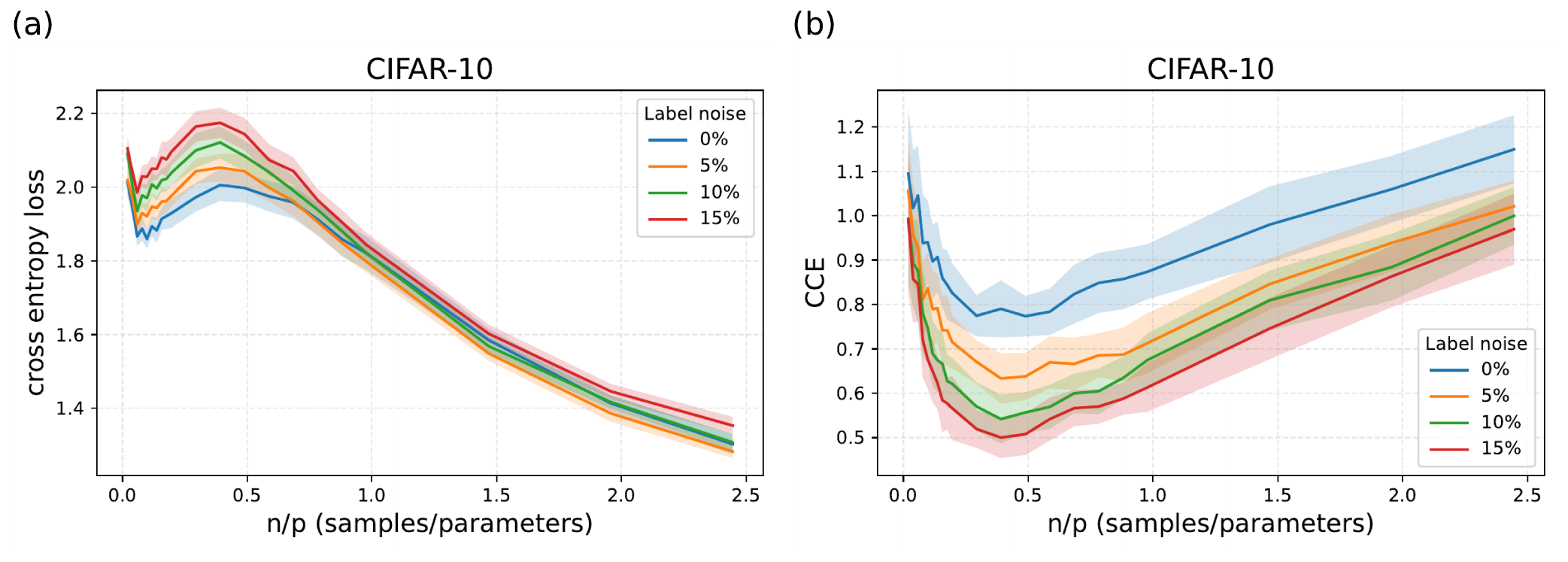}
     \vspace{-8pt}
\caption{\textbf{Representational alignment of independently trained classification networks}: Panel (a) presents the generalization error of the trained networks for various label noise. Panel (b) presents the CCE between penultimate representations of identical networks trained on independent datasets, starting from the same initialization. All curves are plotted as a function of the ratio between the training sample size and the number of network parameters. }
    \label{alignment_classification}
     \vspace{-15pt}
\end{figure}
Finally, we evaluate whether our findings extend to real-world classification tasks. 
We consider the CIFAR-10 object classification task~\cite{krizhevsky2009learning}, using a standard convolutional neural network (CNN), and train two identical networks independently using the same training protocol on non-overlapping subsets of the training set of size $n$, with $n$ ranging from $10^2$ to half of the full training set size (see Sec.~\ref{sec:classification_setup} for details). The CNN is configured with 40 feature maps in the last convolutional layer. We artificially introduce label noise by independently replacing each training label with a random label with probability $p$, where $p$ denotes the label noise rate (e.g., $p=0.1$ corresponds to $10\%$ random labels). The noise is applied once and fixed throughout training, while test labels remain clean. In this case the label noise plays the role of inverse SNR.

Panels~(a) of Fig.~\ref{alignment_classification} show the generalization error as a function of $\gamma = n/p$ for different levels of label noise. We observe the classical double descent pattern. As the label noise increases, the peak in the generalization error becomes more pronounced~\cite{nakkiran2021deep}, consistently with what was shown above for synthetic data. Notably, a peak is also present even at $0\%$ label noise, which may be due to implicit label noise in the ground-truth labels~\cite{northcutt2021pervasive}. The network's performance resembles (qualitatively) the behavior observed in a regression task. Next, we move to the learned representations. Panels~(b) of Fig.~\ref{alignment_classification}, we present the CCE between the penultimate representations as a function of $\gamma$ for different label noise. We observed that the penultimate representations learned with small label noise align better than those learned with higher label noise, in both underparameterized and overparameterized regimes. Similarly, the alignment varies non-monotonically as a function of $\gamma$, with a minimum near the interpolation threshold, akin to what observe for synthetic data. Qualitatively, we observed similar results for the MNIST digit
recognition (\cite{lecun1998mnist}) using a fully connected neural network (see Sec.~\ref{sec.mnist_result} of the Appendix).

\section{Conclusion and Discussion}
\label{sec_conclusion}

We presented a theoretical and computational analysis aimed at studying the alignment between the latent representations in neural networks. To quantify the alignment, we use the Conditional Copula Entropy (CCE), a statistics that in some simple but relevant reference cases can be estimated analytically. The picture which emerges from our analysis is that the CCE behaves qualitatively in the same manner in all the neural networks we considered, including the linear network: the CCE increases monotonically with the signal-to-noise (SNR) ratio, and varies non-monotonically as a function of training sample size, attaining a minimum near the interpolation threshold (see Fig.~\ref{alignment_dln}, Fig.~\ref{alignment_drn}, Fig.~\ref{alignment_dln_drn}, and Fig.~\ref{alignment_classification}).

We further find that different subgroups of networks can achieve the same performance (measured by the generalization error) while exhibiting different degrees of intragroup representational alignment (see Fig.~\ref{alignment_dln} and Fig.~\ref{alignment_drn}). In contrast, a group of networks can outperform another group in a task while showing lower internal alignment than the worst-performing group. This result is consistent with recent work documenting the dissociation between representations and outputs behavior~\cite{lampinen2024learned,braun2025not, lampinen2025representation}.  

 Morever, we show that a network can achieve higher task performance, and yet learn less informative internal representations than a network with lower performance (see Fig.~\ref{alignment_dln_drn}); such differences in representational informativeness (quality) cannot be detected by symmetric measures such as CKA.
This underscores the value of asymmetric alignment measures~\cite{bansal2021revisiting, sucholutsky2025getting,umar2026effect,acevedo2025quantitative}.
This mismatch between generalization performance and representational informativeness highlights that task performance alone is an insufficient proxy for representation quality. A related finding was reported in the context of self-supervised learning, where non-linear autoencoders can produce higher reconstruction loss, yet learn better, more complete representations compared to linear methods with lower reconstruction loss~\cite{mendes2026solvable}.
 
 All together, our results contribute to emerging cross-disciplinary perspectives on representational alignment~\cite{sucholutsky2025getting, klabunde2025similarity} by identifying two key factors that govern alignment — SNR and training sample size — and by providing the first analytic prediction of how these factors shape alignment in a controlled setting.
 
\paragraph{Limitations and Future Directions.}
Our analysis is currently restricted to single-modality settings. However, representational convergence has also been observed across networks trained on different modalities, such as vision and language~\cite{huh2024platonic, acevedo2025quantitative, groger2026revisiting}. Extending our framework to the multimodal setting is a natural and important direction for future work. Additionally, generalizing the theoretical analysis from a linear to a nonlinear setting would be valuable, as it may reveal whether the universality of latent representation behavior we observe empirically is theoretically grounded. Also, it might further expose the influence of network depth and hidden layer width -- factors that are absent from our linear setting results. Finally, studying the dynamic of representational alignment over the course of training is another promising direction. Despite these limitations, our work provides nontrivial insights into the structure of learned representations across different learning regimes and provides a theoretical foundation for future investigations into the origins of convergence of representations.



\section*{Acknowledgments}
We thank Jean Barbier and Sebastian Goldt for their valuable comments and suggestions on the manuscript, and Franky Kevin Nando Tezoh for insightful discussions. A.L. acknowledges financial support from the Friuli Venezia Giulia region (project F53C22001770002).

\bibliographystyle{apalike} 
\bibliography{references}






\newpage
\appendix



\section{Related Work}
\label{sec_related_work}

\subsection{Measures of representational similarity/alignment.}
Quantifying when two information processing systems represent information in compatible ways is a long-standing question, originating in cognitive neuroscience with Representational Similarity Analysis~\cite{kriegeskorte2008representational}, which compares systems through their pairwise dissimilarity matrices rather than raw activations. Modern measures developed for deep networks — Singular Vector Canonical Correlation Analysis (SVCCA)~\cite{raghu2017svcca}, Centered Kernel Alignment (CKA)~\cite{pmlr-v97-kornblith19a}, and shape metrics based on Procrustes distances~\cite{williams2021generalized} — extend this idea while imposing different invariance assumptions (orthogonal, affine, or invertible linear). Recent work has documented systematic biases in these measurements: CKA in particular is sensitive to the feature-to-sample ratio and can report spurious similarity in the regimes we study~\cite{murphy2024correcting}, and finite-neuron sampling distorts both SVCCA and CKA in predictable ways~\cite{kang2025spectral}.  A 
complementary perspective shows that many of these measures can be recast as comparisons of optimal linear readouts~\cite{lampinen2025representation}, clarifying their scope. Crucially, none of these measures is well-suited to 
representations lying on highly curved manifolds~\cite{ansuini2019intrinsic}, motivating our use of a rank-based, information-theoretic alternative in the 
form of the CCE.

\subsection{Convergence of learned representations.} A growing body of empirical work shows that independently trained networks often arrive at structurally similar representations~\cite{li2015convergent,raghu2017svcca,pmlr-v97-kornblith19a,bansal2021revisiting}, a phenomenon recently described as the Platonic Representation Hypothesis~\cite{huh2024platonic}, which conjectures that independently trained deep networks—despite being trained with different objectives and on different data—tend to learn aligned/similar internal representations that converge to a universal statistical model of reality. A subsequent re-evaluation~\cite{groger2026revisiting} argues that, after appropriate calibration, this convergence is most robustly visible in local neighborhood structure rather than in global geometry, precisely the regime our CCE statistic is designed to probe.  

\section{About CCE}
\label{sec_about_CCE}
\subsection{Conditional Copula Entropy}
\label{sec_CCE_definition}
The information content between the pairwise distance variables ($d_x$ and $d_y$) of the representation spaces $x$ and $y$ is measured by the mutual information $\text{I}(d_x, d_y)$. However, mutual information will not be useful in quantifying how well a certain representation space predicts a geometric property of another representation space due to its symmetric nature. \citeauthor{del2024robust} proposed an asymmetric information-theoretic quantity called restricted mutual information, which is defined as 
 \begin{equation*}
\text{I}^\epsilon(d_x \rightarrow d_y) = \int_0^\eta\int_0^\infty dd_xdd_y p(d_x, d_y) \log\left(\frac{p(d_x, d_y)}{p(d_x)p(d_y)}\right) 
\end{equation*} 
where $\eta = F^{-1}_x(\epsilon)$ and $F_x(r) = \int_0^r p(d_x) dd_x$ denote the cumulative distribution function of $d_x$. This quantity is proposed based on the intuition that a good distance space should be locally semantically meaningful. i.e, a space where a data point is enclosed by points that are "similar". Therefore, in the limit of small $\epsilon$, the restricted mutual information quantifies the shared information by the random variables $d_x$ and $d_y$ when conditioned on the events $d_x = \eta $. With the assumption that large distances are not informative, the authors (\citeauthor{del2024robust}) interpret this measure as the actual information in $d_y$ that is contained in $d_x$. By expanding the restricted mutual information in the limit $\epsilon \rightarrow 0$, the first nonzero term is its first derivative around zero:
\begin{align*}
  \lim_{\epsilon \rightarrow 0} \frac{\text{I}^\epsilon(d_x \rightarrow d_y)}{\epsilon}&=\frac{\partial}{\partial \epsilon}\text{I}^\epsilon(d_x; d_y)|_\epsilon\\&=\frac{1}{p(d_x = \eta )}\frac{\partial}{\partial d_x}\text{I}^\epsilon(d_x; d_y)|_\eta \\ &= \frac{1}{p(d_x = \eta  )}\frac{\partial}{\partial d_x}\bigg[\int_0^c\int_0^\infty dd_x dd_y p(d_x, d_y) \log\left(\frac{p(d_x, d_y)}{p(d_x)p(d_y)}\right)\bigg] \bigg|_\eta 
  \\&= \frac{1}{p(d_x = \eta )}\frac{\partial}{\partial d_x}\int_0^c dd_x p(d_x) \bigg[ \int_0^\infty dd_y p(d_y| d_x) \log\left(\frac{p(d_y|d_x)}{p(d_y)}\right)  \bigg]\bigg|_\eta \\
&=  \lim_{\eta  \rightarrow 0}\int_0^\infty dd_y p(d_y| d_x=\eta ) \log\left(\frac{p(d_y|d_x=\eta )}{p(d_y)}\right)
\end{align*}
Therefore, in the limit of $\epsilon \rightarrow 0$, the restricted mutual information is given by
\begin{equation}
     \lim_{\epsilon \rightarrow 0} \frac{\text{I}^\epsilon(d_x \rightarrow d_y)}{\epsilon} = \lim_{\eta  \rightarrow 0} D_{\mathrm{KL}}[p(d_y| d_x = \eta ) \parallel p(d_y)],
\end{equation}
where $D_{\mathrm{KL}}$ denotes the Kullback-Leibler divergence (a pseudo-distance). This divergence represents the amount of information (in nats) gained about $d_y$ when conditioning on $d_x = \eta$, compared to the marginal distribution of $d_y$. It quantifies how much the distribution of $d_y$ changes when conditioned on small distances in space $x$—the greater the change, the more informative space $x$ is with respect to space $y$.
Let 
\begin{equation*}
    c = (c_x, c_y) = (F_x(d_x), F_y(d_y)), 
\end{equation*}
where $F_x(d_x)$ and $F_y(d_y)$ are the cumulative densities of random variables $d_x$ and $d_y $. According to Sklar's theorem \cite{nelsen2006introduction}, the distribution of $d = (d_x, d_y)$ can be decomposed as the product of the marginals and the joint copula distribution:
\begin{equation*}
    P(d_x, d_y) = p_c(c_x, c_y) p(d_x)p(d_y),
\end{equation*}
where $p_c(c_x, c_y)$ is the joint copula distribution. This relationship between the distribution of $c$ and its respective copula distribution allows us to rewrite the restricted mutual information as
\begin{align}\label{rmi}
   \lim_{\epsilon \rightarrow 0} \frac{\text{I}^\epsilon(d_x \rightarrow d_y)}{\epsilon}&= \lim_{\eta \rightarrow 0} \int_0^\infty dd_y p(d_y\;|\; d_x =\eta) \log\left(\frac{p(d_y \;|\; d_x=\eta)}{p(d_y)}\right)\\
   &= \lim_{\epsilon \rightarrow 0} \int_0^1 d c_y p_c(c_y\;|\; c_x =\epsilon) \log p_c(c_y \;|\; c_x=\epsilon)\\
   &= -\lim_{\epsilon \rightarrow 0} H(c_y | c_x = \epsilon).
\end{align}
\subsection{Properties of CCE}
\label{sec.CCE_properties}
Importantly, CCE (similarly, II) satisfies the standard invariance properties expected from the representation alignment measures~\citep{pmlr-v97-kornblith19a}. In particular, it is invariant to isotropic scaling and orthogonal transformations, as it depends only on the distribution of pairwise distances. Formally, for any $\alpha, \beta \in \mathbb{R}$ and orthogonal matrices $U, V$,
\begin{equation*}
    \text{CCE}(\alpha A \to \beta B) = \text{CCE}(A \to B), \quad
    \text{CCE}(AU \to BV) = \text{CCE}(A \to B).
\end{equation*}

\section{Deep Linear Network Gradient Flow Equations}
For a deep linear network with $D$ hidden layers, continuous-time gradient descent on the mean squared training error yields the following dynamics for each weight matrix $W_l$:
\begin{equation}
    \tau \frac{dW_l}{dt} = \bigg(\prod_{i= l+1}^D W_i\bigg)^T \bigg[yX - \bigg(\prod_{i=1}^D W_i\bigg)X^TX\bigg] \bigg(\prod_{i=1}^{l-1} W_i\bigg)^T,
\end{equation}
where $\tau = n/2$ is a time constant inversely proportional to the learning rate. These dynamics capture the behavior of gradient descent in the limit of a small learning rate. 

\citet{advani2020high} provides a detailed description of these dynamics starting from small random initializations, a setting commonly referred to as the rich learning regime \cite{woodworth2020kernel}. Based on the observation that gradient descent dynamics in deep linear networks initialized with small random weights extract low rank structure \cite{saxe2013exact}, they begin by using a singular value decomposition (SVD) based change of variables. Specifically, they define
\begin{align*}
    W_1 &= r_2 z(t) V^T, \\
    W_l &= c(t) r_{l+1} r_l^T, \quad \text{for } l = 2, \dots, D,
\end{align*}
where $V$ is the matrix of right singular vectors of $X$, $z(t) \in \mathbb{R}^{1 \times d}$ is a vector encoding the time-varying overlap with each principal axis in the input, $r_l \in \mathbb{R}^{d_l \times 1}$ are arbitrary unit-norm vectors that capture the freedom in the network's internal representations, and $c(t)$ is a scalar which encodes the change in representation over time. With these definitions, the total weight matrix $W_{\text{tot}} = \prod_{i=1}^D W_i$ simplifies to
\begin{equation*}
    W_{\text{tot}} = \bigg( \prod_{i=2}^D c(t) r_{i+1} r_i^T \bigg) r_2 z(t) V^T = u(t) z(t) V^T,
\end{equation*}
where $u(t) = c(t)^{D-1}$ and we set $r_{D+1} = 1$.

Further decomposing the input correlation matrix $X^TX$ and the input–output correlation matrix $yX$ as
\begin{equation*}
    X^TX = V \Lambda V^T, \qquad yX = \tilde{s} V^T,
\end{equation*}
with $\tilde{s} = z^* \Lambda + \tilde{\epsilon} \Lambda^{1/2}$ and $z^* = w^* V$, the network dynamics reduce to the following system of equations:
\begin{align}
    \tau \frac{d u(t)}{dt} &= u(t)^{\frac{2D-4}{D-1}} \bigg( \tilde{s} z(t)^T - u(t) z(t) \Lambda z(t)^T \bigg), \label{eq:u_dynamics}
    \\
    \tau \frac{d z(t)}{dt} &= u(t) \bigg( \tilde{s} - u(t) z(t) \Lambda \bigg).
    \label{eq:z_dynamics}
\end{align}
Or in component-wise notation
\begin{align}
    \tau \frac{d u(t)}{dt} &=   u(t)^{\frac{2D-4}{D-1}} \sum_{i=1}^d\bigg( \lambda_i z^*_iz_i(t) - u(t)\lambda_i z^2_i(t) + \lambda^{1/2}\tilde{\epsilon}_i z_i(t)\bigg),
    \\
    \tau \frac{d z_i(t)}{dt} & = u(t)\bigg(\lambda_i z^*_i - u(t) \lambda_i z_i(t) + \lambda^{1/2}\tilde{\epsilon}_i\bigg) \quad \text{ for }\; i = 1,... ,d.
    \end{align}
These equations represent a substantial reduction in complexity. While full gradient descent in a $D$-layer network with $d$ hidden units per layer involves  $O(d^2 D)$ parameters, the reduced system requires only $d+1$ parameters, irrespective of depth and hidden layer width. Despite this dramatic compression, \citet{advani2020high} demonstrates that the reduced dynamics provide an accurate description of the full training trajectory, provided the network is initialized with sufficiently small random weights.

\subsection{Gradient Flow fixed points}
For a two-layer linear neural network, i.e., $D=2$, the gradient flow training dynamic on the mean squared training error yields the following dynamics for $W_1$ and $W_2$:
\begin{equation}
\label{gradient_flow_W1}
    \tau \frac{dW_1}{dt} = W_2^T \bigg(yX -  W_2W_1X^TX\bigg),
\end{equation}
\begin{equation}
\label{gradient_flow_W2}
    \tau \frac{dW_2}{dt} =\bigg(yX -  W_2W_1X^TX\bigg) W_1^T,
\end{equation}
These dynamics of the weight contain two manifolds of  fixed points: An unstable fixed point  at zero $\mathcal{U}$ and a stable fixed points at the global minimum $\mathcal{S}$, which are given by
\begin{align*}
    \mathcal{U} &= \big\{ W_2 = 0, W_1 = 0\big\}\\
    \mathcal{S} &= \big\{ W_2, W_1 \;|\; W_2W_1X^TX = YX\big\}. 
\end{align*}

Using the reduced network dynamics in equations 
\ref{eq:z_dynamics}, \ref{eq:u_dynamics} of $z$ and $u$
\begin{align}
    \tau \frac{d u(t)}{dt} &= \bigg( \tilde{s} z(t)^T - u(t) z(t) \Lambda z(t)^T \bigg),
    \\
    \tau \frac{d z(t)}{dt} &= u(t) \bigg( \tilde{s} - u(t) z(t) \Lambda \bigg).
\end{align}
Let $v(t) = u(t)z(t)$. Then
\begin{align*}
    \tau \frac{dv(t)}{dt} &= \tau \left[ u(t) \frac{dz(t)}{dt} + z(t) \frac{du(t)}{dt} \right]\\
     &= \tau \left[ u(t)^2 \tilde{s} - u(t)^3 \Lambda z(t) + \tilde{s} z(t) z(t)^\top z(t) - u(t) \Lambda z(t) z^\top z(t) \right] \\
     &= \tau \left[ \tilde{s} \left( u(t)^2 I + z(t) z(t)^\top \right) - u(t) \Lambda \left( u(t)^2 I + z(t) z(t)^\top \right) z(t) \right] \\
    &= \tau \big(\tilde{s} - v(t) \Lambda\big) \big(u(t)^2I + z(t)^\top z(t)\big).
\end{align*}
Because the matrix $u(t)^2 I + z(t)^\top z(t)$, which depends on $z(t)$ and $u(t)$ in a nonlinear way, the dynamics remain coupled. As a result, a closed-form solution for the finite time is not feasible. However, at the global minimum / stable fixed point (i.e., when $t \to \infty$), we must have
\begin{equation*}
    \big(\tilde{s} - v^* \Lambda\big) \big(u^{*2}I + z^{*\top} z^*\big) = 0
\end{equation*}
\begin{equation*}
    \big( w^* V \Lambda + \tilde{\epsilon} \Lambda^{1/2} - v^* \Lambda \big)\big(u^{*2}I + z^{*\top} z^*\big) = 0,
\end{equation*}
where $\Lambda$ is the diagonal matrix of eigenvalues of $X^\top X$ and $\tilde{\epsilon} \sim \mathcal{N}(0, \sigma_\epsilon I_d)$. Since $u^{*2}I + z^{*\top} z^*$ is always positive definite, this implies that at the stable fixed point
\begin{equation}
\label{v_star}
    v^* = \big( w^* V \Lambda + \tilde{\epsilon} \Lambda^{1/2} \big)\Lambda^{-1}.
\end{equation}

\section{Exact Generalization Error of Two-layer Neural Network}
\label{exact_gen_error_dln}
As shown in Ref~\cite{advani2020high}, the generalization error of the student network at the global minimum is
\begin{equation}
\label{gr}
   \mathcal{E}_g := \mathbb{E}_{w^*, x,\epsilon}[(y - \hat{y})^2] = \frac{1}{d}\mathbb{E}_{w^*, v}[(w^*V^\top- v)^2] + \sigma_\epsilon^2,
\end{equation} 
where the expectation is over all potential teacher weight $w^*$, input $x$ and noise $\epsilon$. Substituting
the global minimum in Eq.~\ref{gr} yields
\begin{equation}
\label{egr}
    \mathcal{E}_g = \sigma_w^2 + \sigma_\epsilon^2
    - \frac{1}{d}\sum_{i=1}^{\operatorname{rank}(X^\top X)}
      \!\left(\sigma_w^2 - \frac{\sigma_\epsilon^2}{\lambda_i}\right),
\end{equation}
where $\{\lambda_i\}$ are the non-zero eigenvalues of $X^\top X$, and $\sigma_w^2$, $\sigma_\epsilon^2$ are the variance of the teacher weight and the noise, respectively. Eq.~\ref{egr} reveals the critical role of the eigenvalue spectrum of the input covariance matrix. Small eigenvalues degrade generalization, while large eigenvalues improve it. Directions with $\lambda_i= 0$ are never learned: the corresponding components $v_i$ remain at their initialized values throughout training \citep{advani2020high}, which is why the sum in Eq.~\ref{egr} runs only over directions with non-zero eigenvalues. 
Consequently, initialization directly affects generalization: if weights are initialized to small values, the contribution of these frozen directions to the output remains negligible. 

In the proportional high-dimensional limit $n, d \to \infty$ with $\alpha = n/d$ fixed, the empirical spectral distribution of $X^\top X$ converges to the
Marchenko--Pastur law~\citep{marvcenko1967distribution},
\begin{equation}
    \label{sup_mp_distn}
    f(\lambda)
    = \frac{\sqrt{(\lambda_{+} - \lambda)(\lambda - \lambda_{-})}}{2\pi\lambda}
    + (1 - \alpha)^{+}\,\delta(\lambda),
\end{equation}
where $\lambda_{\pm} = (1 \pm \sqrt{\alpha})^{2}$ and
$(\cdot)^{+} = \max(\cdot,\, 0)$.
The first term is the continuous bulk on
$[\lambda_{-}, \lambda_{+}]$; the second captures the point mass at zero when
$\alpha < 1$. 
Replacing the empirical average with an integral over $f$, 
the generalization error admits the asymptotic form
\begin{equation}
    \label{generalization_error}
    \mathcal{E}_g
    = \sigma_w^2 + \sigma_\epsilon^2
    - \int\!\left(\sigma_w^2 - \frac{\sigma_\epsilon^2}{\lambda}\right)f(\lambda)\,d\lambda.
\end{equation}
\section{Asymptotic Alignment}
\label{asymptotic_alignment_proof}
 Suppose that we trained two linear networks $\hat{y}^a_i = W_2^a W_1^a x_i$ and $\hat{y}^b_i = W_2^b W_1^b x_i$ by gradient descent from small random initializations on independent datasets $\mathcal{D}^a = \{X^a, y^a\}$ and $\mathcal{D}^b = \{X^b, y^b\}$, drawn independently from the same data distribution. 
Using the singular value decomposition (SVD) parameterization, and assuming that the correlation matrices share the same orthogonal singular vector matrix $V$ — as is commonly assumed in theoretical analysis of deep linear networks \cite{saxe2013exact,advani2020high,domine2024lazy,jarvis2025theory} — we express the first-layer weights as
\begin{equation*}
W_1^a=r^a z^aV^T = r^a \tilde{z}^a,
\qquad
W_1^b = r^bz^bV^T= r^b \tilde{z}^b.
\end{equation*}
This parameterization assumes that the first-layer weights have a low-rank structure. It is a well-known result from prior work that deep linear networks tend to extract low-rank structures when initialized with small weights \cite{arora2019implicit,saxe2013exact}, which is a form of implicit inductive bias by the training algorithm \cite{kalimeris2019sgd,refinetti2023neural}.
We notice that for any test inputs $x$ and $x'$, the distance between them in the projected space by $W_1^a$ is
\begin{equation}
\left\|
r^a \tilde{z}^ax
-
r^a \tilde{z}^ax'
\right\|^2
=
||r^a||^2
\left\|
\tilde{z}^a(x-x')
\right\|^2=
\left\|
\tilde{z}^a(x-x')
\right\|^2.
\end{equation}
Analogously, in the projected space defined by $W_1^b$.
Let \(A = W_1^a X_{\mathrm{test}}^\top\) and \(B = W_1^b X_{\mathrm{test}}^\top\). Then, the representations \(A\) and \(B\) are equivalent, in terms of pairwise distances, to one-dimensional projections
$\tilde{A} = \tilde{z}^a X_{\mathrm{test}}^\top$ and 
$\tilde{B} = \tilde{z}^b X_{\mathrm{test}}^\top$. As a consequence, the CCE between $A$ and $B$ is the same as the CCE between $\tilde{A}$ and $\tilde{B}$.

In Ref.~\cite{umar2026effect}, it is shown that if $x \sim \mathcal{N}(0, I_d)$, and define $U = u^\top x$ and $V = v^\top x$, where $u, v \in \mathbb{R}^d$. Then $U$ and $V$ are jointly Gaussian, and the CCE admits the following closed-form expression
\begin{equation} 
\label{CCE_form_a}
    \text{CCE}(U \to V) = \text{CCE}(V \to U) =-\frac{1}{2}\log(1 - \rho^2) - \frac{1}{2}\rho^2,
\end{equation}
where $\rho = \mathbb{E}\big[u^\top v\big]/
\sqrt{
\mathbb{E}\big[\|u\|^2\big]\;
\mathbb{E}\big[\|v\|^2\big]
}$ denotes the correlation coefficient between $U$ and $V$.
Therefore, the CCE of $A$ and $B$ depends only on the correlation $\rho$ between $\tilde{A}$ and $\tilde{B}$.
\begin{equation}
\label{rho_sup}
    \rho
=
\frac{
\mathbb{E}\big[z^a z^{b\top}\big]
}{
\sqrt{
\mathbb{E}\big[\|z^a\|^2\big]\;
\mathbb{E}\big[\|z^b\|^2\big]
}
}.
\end{equation}
\subsection{Correlation between A and B}
At convergence, $z^{*a}$ takes the form
\begin{equation}
z^{*a} = \frac{1}{u^*}\!\left(w^* V + \tilde{\epsilon}^a (\Lambda^a)^{-1/2}\right),
\end{equation}
where $\Lambda^a$ is the diagonal matrix of eigenvalues of $X^{a\top} X^a$, $\tilde{\epsilon}^a \sim \mathcal{N}(0, \sigma_\epsilon^2 I_d)$, and $u^*$ is the magnitude of the learned weight. An analogous expression holds for $z^{*b}$. Since both networks align with the same principal directions \(V\), their features concentrate along the signal direction and $\rho(v^{a*}, v^{b*})=\rho(u^*z^{a*}, u^*z^{b*})=\rho(z^{a*}, z^{b*})$ by the fact that correlation is invariant to global scaling. Substituting the asymptotic solutions into $\rho$ (Eq.~\ref{rho_sup}) yields the following expression.
\begin{equation*} 
\rho^*
=
\frac{
\mathbb{E}\big[z^{*a} z^{*b\top}\big]
}{
\sqrt{
\mathbb{E}\big[\|z^{*a}\|^2\big]\;
\mathbb{E}\big[\|z^{*b}\|^2\big]
}
}
= \frac{\sum_{i=1}^{R} \sigma_w^2}{\sqrt{\bigg[\sum_{i=1}^{R} \left(\sigma_w^2 + \frac{\sigma_\epsilon^2}{\lambda_i^a}\right)\bigg] \bigg[\sum_{i=1}^{R} \left(\sigma_w^2 + \frac{\sigma_\epsilon^2}{\lambda_i^b}\right)\bigg]}},
\end{equation*}
where $\{\lambda_i^a\}$ and $\{\lambda_i^b\}$ are the non-zero eigenvalues of $X^{a\top} X^a$ and $X^{b\top} X^b$, respectively, $R$ is the rank of the input covariance matrix, and $\sigma_w^2$, $\sigma_\epsilon^2$ are the variances of the teacher weights and noise. Replacing the empirical average over eigenvalues with an integral against $f$ (Eq.~\ref{sup_mp_distn}) yields
 
\begin{equation}
    \rho^* =
    \begin{cases}
    \displaystyle
    \frac{\text{SNR}}{\text{SNR}+ \frac{1}{\alpha - 1}}, & \alpha > 1, \\[10pt]
    \displaystyle
    \frac{\alpha \cdot \text{SNR}}{\text{SNR} + \frac{1}{1 - \alpha}}, & \alpha < 1,
    \end{cases}
    \end{equation}
    where $\rho^* = \lim_{t \to \infty} \rho(t)$, $\mathrm{SNR} = \sigma_w^2 / \sigma_\epsilon^2$, and $\alpha = n/d$ denotes the ratio of training samples to input dimension.

\section{Classification Task}
\subsection{Experimental Setup}
\label{sec:classification_setup}
We performed experiments on a two-baseline neural network for a classification task: A fully connected neural network (FCNN) and a standard convolutional neural network (CNN). The two networks have the following structure:
\begin{enumerate}
    \item 
A single-layer FCNN with $k$ hidden units, with input and output dimensions fixed to match the number of pixels in the input image and the number of classes in the task, respectively. 
\item  A family of  CNN consisting of $4$ convolutional stages of width $ [k, 2k, 4k, 8k]$, where $k$ is the width parameter, followed by a fully connected layer as a classifier. The Maxpoll is $[2,2,2,4]$.  For the entire convolution layer, the kernel size $=3$, stride $=1$, and padding $=1$. This architecture is identical to the one considered in \cite{gu2024unraveling} 
\end{enumerate}
The networks are trained using the mini-batch SDG algorithm to minimize the cross entropy loss. A batch size of $128$ samples is used.  For FCNNs, we set the learning rate to start with $l_0=0.1$ and after every successive $50$ epoch it changes to $l_r = l_0/(\sqrt{1 + epoch/50})$. While for CNNs, the learning rate starts with $l_0=0.05$, and updates as $l_r = l_0/(\sqrt{1 + epoch\times50})$ after every $50$ epoch. This learning rate scheme was used in a recent paper by \cite{gu2024unraveling}.
The FCNN is trained for image classification tasks with the MNIST dataset, and the CNN is trained for image classification tasks with the CIFAR-10 dataset. 
For both datasets, we partition all the training data into two equal part. i.e., we have two sets of training data, for MNIST each set has $3 \times 10^4$ samples. While for CIFAR-10, each set has $25\times 10^3$ samples. 
\subsection{Preliminary Results on Mnist}
 \label{sec.mnist_result}
\begin{figure}[h]
    \centering
    \includegraphics[width=1\linewidth]{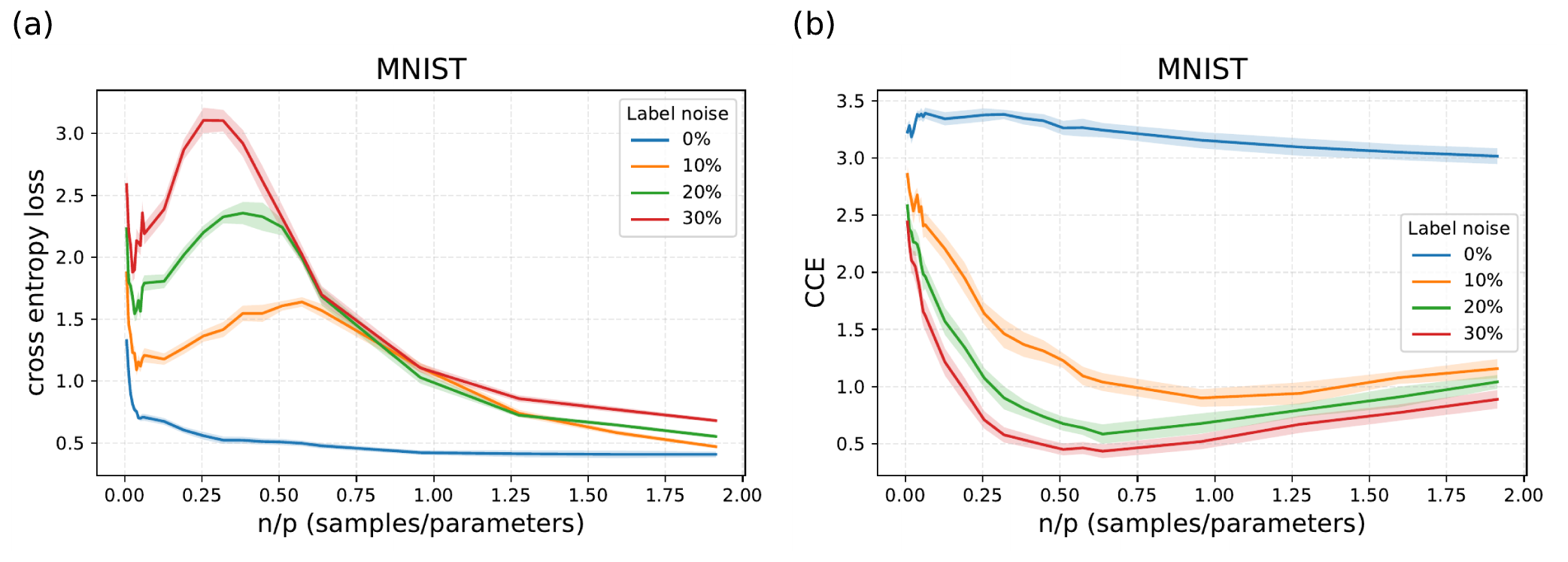}
\caption{\textbf{Representational alignment of independently trained classification networks}: Panel (a) presents the generalization error of the trained networks for various label noise. Panel (b) presents the CCE between penultimate representations of identical networks trained on independent datasets, starting from the same initialization. All curves are plotted as a function of the ratio between the training sample size and the number of network parameters. }
    \label{alignment_classification_mnist}
\end{figure}

\newpage

\end{document}